\documentclass[a4paper,twoside]{article}

\usepackage{epsfig}
\usepackage{subfigure}
\usepackage{calc}
\usepackage{amssymb}
\usepackage{amstext}
\usepackage{amsmath}
\usepackage{amsthm}
\usepackage{multicol}
\usepackage{pslatex}
\usepackage{ICAIIT}   

\usepackage{xcolor}
\usepackage{color}
\usepackage{tabularray}

\usepackage{graphicx}
\usepackage{amsmath}
\usepackage{multirow}
\usepackage{changepage}
\usepackage{footnote}
\usepackage[marginal]{footmisc}

\begin{document}

\title{Multi-view Multi-label Anomaly Network Traffic Classification based on MLP-Mixer Neural Network}

\author{\authorname{Yu~Zheng\sup{1}, Zhangxuan~Dang\sup{1},Chunlei~Peng\sup{1},Chao~Yang\sup{1},Xinbo~Gao\sup{2}}
\affiliation{\sup{1} Xidian University}
\affiliation{\sup{2}Chongqing University of Posts and Telecommunications}
}

\abstract{Network traffic classification is the basis of many network security applications and has attracted enough attention in the field of cyberspace security. Existing network traffic classification based on convolutional neural networks (CNNs) often emphasizes local patterns of traffic data while ignoring global information associations. In this paper, we propose an MLP-Mixer based multi-view multi-label neural network for network traffic classification. Compared with the existing CNN-based methods, our method adopts the MLP-Mixer structure, which is more in line with the structure of the packet than the conventional convolution operation. In our method, one packet is divided into the packet header and the packet body, together with the flow features of the packet as input from different views. We utilize a multi-label setting to learn different scenarios simultaneously to improve the classification performance by exploiting the correlations between different scenarios. Taking advantage of the above characteristics, we propose an end-to-end network traffic classification method. We conduct experiments on three public datasets, and the experimental results show that our method can achieve superior performance.}

\onecolumn \maketitle \normalsize \vfill

\section{Introduction}
\label{sec:introduction}
{W}{ith} the development of the Internet and the popularization of mobile Internet devices, the network environment has become extremely complex. For Internet providers and network administrators, how to provide secure and reliable network services is a crucial issue. In particular, identifying anomaly network traffic from the massive normal network traffic is the basis for providing satisfactory network services. Network anomaly detection aims to identify malicious traffic to prevent network intrusion and provide network security protection. Therefore, network traffic classification is an essential technology for network anomaly detection. In addition, network traffic classification can also help rationally allocate network resources and guarantee quality-of-service (QoS)~\cite{dainotti2012issues}. By associating network traffic with traffic-generating applications or application types, different solutions can be provided in different situations.

The existing network traffic classification methods can be divided into three categories: port-based, deep packets inspection (DPI)-based, and machine learning-based. The port-based classification method partitions the application of generating the packet according to the port number in the packet header. But as the increasing types of network traffic, the utilizing of random ports, and the prevailing of peer-to-peer (P2P), this method has become almost unreliable~\cite{barut2020netml}. DPI examines packet content to perform more complex syntax matching to classify the application of the packet. However, due to the popularity of network protocol encryption technology and data packet tunneling technology, this method can only achieve a low accuracy~\cite{dainotti2012issues}. Machine learning methods classify packets by analyzing and matching patterns of selected features. But the feature extraction is often the bottleneck for the machine learning method. Good feature selection and representation will greatly improve the performance of machine learning methods. Feature design often requires a lot of domain knowledge, and manual feature extraction is likely to ignore underlying interrelationships between different features. In addition, feature extraction methods can also become outdated with changing in traffic conditions~\cite{holland2021new}.

\begin{figure}[t]
\begin{center}
\includegraphics[width=0.9\linewidth]{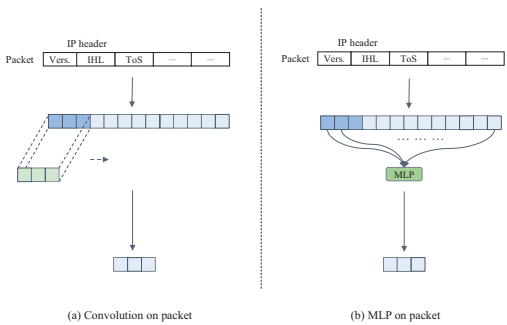}
\end{center}
   \caption{\emph{For the same data packet, convolution focuses on local patterns, while MLP can pay more attention to the global information in the protocol structure of the data packet.}}
\label{fig:FLS}
\end{figure}

In recent years, with the popularity of deep learning technology, deep neural network has achieved excellent results in many research fields, such as: image classification~\cite{chen2022label}, speech recognition~\cite{hinton2012deep} and network security~\cite{javaid2016deep,xu2020method,rahman2020mockingbird}. Compared with conventional machine learning methods, deep learning has better nonlinear mapping ability and can learn more complex data. In terms of feature extraction, deep learning can extract features from raw data for learning under the premise of reducing manual intervention. Specially, deep learning has also been widely studied in network traffic classification~\cite{wang2017malware,soleymanpour2021cscnn,lotfollahi2020deep,kim2018multimodal}.

However, existing deep learning-based network traffic classification methods often employ convolutional operations to complete feature extraction on the original data packet. Convolution operations often appear in image recognition and detection because objects in the image are characterized by local patterns, and the convolution operations focus on the local features of the image, therefore, convolution operations can often achieve good performance in image-related tasks. For raw data packets, it consists of protocols working at different layers of OSI. The protocols of each layer have completely different structures and semantics with few associated local patterns. Therefore, it is unreasonable to use convolution operations to extract local patterns of the semantic information in the protocol, and it is likely to ignore the different information of various protocols, as showed in Fig.~\ref{fig:FLS}. Furthermore, in most network traffic classification studies, the entire packet or specific parts of the packet are usually fed into the classification model, which does not consider the different clues from different parts of the packet.

In this paper, we propose a MLP-Mixer based multi-view multi-label neural network architecture for network  traffic classification. The powerful MLP-Mixer \cite{tolstikhin2021mlp} is employed in our model, which uses the simple MLP to mix different features to obtain useful information. Such structure does not depend on the local patterns of the underlying data, so we can use it instead of the convolutional operations, which often used in conventional deep network traffic classification. MLP-Mixer is used to extract and mix information from different views.

In our model, three different views are utilized to perform feature extraction on the traffic data. The first view focuses on the packet header. The packet header is made up of protocol fields that contain short and efficient information that can fully describe the packet. We hope that the model can automatically extract features on the header with high information concentration. The second view focuses on the body of the data packet. The packet body often contains a variety of upper-layer protocols or encrypted data, and the information in the packet body is complex and obscure. 
We hope that the model can extract as much useful information as possible from this view through the feature representation capabilities of the neural network to improving the performance of the model. Our third view focuses on flow features of the packet. Since it is impossible to completely separate individual packets from the flow to discuss, we set the model to focus on the flow features of the packets.

In the network traffic classification task, different scenarios need to be assigned labels according to the attributes of the traffic, such as assigning labels according to normal traffic and abnormal traffic, or assigning labels according to the activities that the traffic participates in. In these different scenarios, these attributes of the traffic are correlated and can be utilized to learn more discriminitive model. Therefore, our method assigns multi-label to the traffic data. Multi-label is dedicated to getting better feature representation by propagating the correlations from different labels to improve model learning efficiency.

Besides, byte encoding is used to represent packets. Byte encoding is used by many studies of network traffic classification due to its simplicity~\cite{wang2017malware,lotfollahi2020deep}. However, the focus of these studies is usually on the introduction of the model, and the encoding method is rarely mentioned, so we will give a brief introduction to the byte encoding.

To evaluate the performance of the proposed method on network traffic classification, three public network traffic datasets, ISCX VPN and non-VPN~\cite{draper2016characterization}, Tor and non-Tor~\cite{lashkari2017characterization}, USTC-TFC2016~\cite{wang2017malware} are utilized in experiments. Compared with other network traffic classification methods, the contribution of the proposed method is threefold:
\begin{itemize}
\item To the best of our knowledge, this is the first time that the Mixer has been used in network traffic classification. The Mixer helps to better understand the semantics of the byte-encoded underlying data and is more interpretable than convolution operations.
\item Multi-view is used to independently input and extract features from different parts of the packet, which can take into account the differences in the information provided by different parts of the packet.
\item Multi-label framework is used in proposed method. It can deal with the correlation between different labels to solve the multi-scene classification problem, at the same time, improve the performance of the network traffic classification.
\end{itemize}

The rest of this paper is organized as follows. In Section 2, the recent network traffic classification methods and development of the deep networks are presented. In Section 3, our network traffic classification model is introduced in detail. Section 4 includes the experimental results and analysis. A conclusion is mentioned in Section 5.

\section{Related Work}
\subsection{Traffic classification}

Some recent studies on network traffic classification mostly use machine learning or deep learning methods to classify network traffic.

Wang et al.~\cite{wang2017malware} adopted CNN to classify malicious traffic using the USTC-TFC2016 dataset they created. In the paper, the network traffic was divided according to flow or session, and then combined with the original data packets to convert into gray images and feed them to 2D-CNN for classification.
In another work by Wang et al.~\cite{wang2017end}, an end-to-end traffic classification framework was proposed, in which 1D-CNN was used as the classification model. Bu et al.~\cite{bu2020encrypted} proposed two parallel networks to classify the packet header and body respectively, and then weighted the results of the two networks to obtain the final classification result. At the same time, the network-in-network structure~\cite{lin2013network} was combined into 1D-CNN in the classification network, which enhanced the abstraction ability of the model.
Lashkari et al.~\cite{habibi2020didarknet} proposed the DeepImage to classify darknet datasets. In this method, trees classifier was used to rank the features of the traffic set, and then the selected features were converted into 2D gray images, and then fed into a simple 2D-CNN for classification.
Lotfollahi et al.~\cite{lotfollahi2020deep} used two different network structures, stacked autoencoder and 1D-CNN, to classify raw packets. The experiments showed that 1D-CNN had better performance. Sarkar et al.~\cite{sarkar2020detection} proposed a DNN model to classify manually extracted features in Tor traffic. Shen et al.~\cite{shen2021accurate} proposed GraphApp. The encrypted  Decentralized Applications (DApps) flow was converted into a graph structure of traffic interaction, and then the DApp was fingerprinted based on the graph neural network. Shen et al.~\cite{shen2020fine} proposed FineWP, which only used the length of the data packet as a feature, and then classified it through conventional machine learning models to achieve good results. Based on existing deep learning techniques, such as CNN, LSTM, etc., Aceto et al.~\cite{aceto2019mobile} designed a mobile traffic classifier that automatically extracted features, and evaluated the classification effects of different model structures.
Draper et al.~\cite{draper2016characterization} studied the effectiveness of flow-based time-correlated features to detect VPN traffic, using two well-known machine learning methods, C4.5 and KNN, to classify them.
Holland et al.~\cite{holland2021new} proposed nPrintML, a tool that can uniformly represent packets and automate model tuning.

\subsection{Network Structure}

The design and innovation of model architecture largely drives the development of network traffic classification. We review some important network structures in the history of model development.

It is widely considered that Lecun et al.~\cite{lecun1989backpropagation} first implemented a CNN model. In the paper, the convolution operation reduced the parameters of the network using the idea of weight sharing and applied backpropagation to the CNN. It had laid a solid foundation for the application of CNN in various fields. Krizhevsky et al.~\cite{krizhevsky2012imagenet} proposed the Alexnet and won the title in 2012 ImageNet ILSVRC by a significant margin. Alexnet stacked multiple convolutional layers and fully connected layers, and used ReLU as an activation function, Dropout and other techniques to demonstrate the great potential of deep learning. Then came deeper neural networks, such as VGG~\cite{simonyan2014very} and GoogleNet~\cite{szegedy2015going}. He et al. \cite{he2016deep} proposed ResNet. The use of building blocks with Shortcut connection in Resnet effectively solved the vanishing gradient problem of deep neural networks.

Vaswani et al.~\cite{vaswani2017attention} proposed a Transformer using an attention mechanism. The Transformer abandoned the traditional CNN and RNN, completely used the attention mechanism, which can reduce any two positions in the sequence to a constant.

In particular, Tolstikhin et al.~\cite{tolstikhin2021mlp} proposed a network structure without convolution and attention mechanism, MLP-Mixer, for classification.  One of the most unique structures is the Mixer layer. Using the kernel in convolutional structures to obtain useful information in values at different locations and in different channels. In MLP-Mixer, this step is divided into two MLP networks for processing. The Mixer layer consists of two MLP networks, which perform channel mixing and token mixing respectively, and are able to communicate the features of different channels and spatial locations. In addition to the MLP structure, the Mixer layer also uses skip-connections and layer normalization to enhance the performance of the network. The MLP-Mixer obtained by stacking Mixer layers approaches the state-of-the-art performance on classification tasks, which demonstrates the great potential of the simple-structured MLP.

\subsection{Multi-label Learning}
The multi-label classification task indicates that one data input to the model will have a series of labels, and these labels often contain some dependencies. Multi-label classification tasks are usually more in line with the property that real-world objects are complex and have multiple semantics. Below we present recent research in multi-label learning.

Zhang et al.~\cite{zhang2018multilabel} proposed a regional latent semantic dependencies model (RLSD). This model used a fully convolutional network to localize semantically relevant regions.This model had a good effect, especially on the prediction of small objects occurring in the images.
Chen et al.~\cite{chen2019multi} proposed a multi-label classification model based on Graph Convolutional Network (GCN). GCN can use graphs to spread information among multiple labels, and to learn the dependencies between different labels. Then a set of interdependent classifiers were obtained to classify the image descriptors extracted by another subnet.
You et al.~\cite{you2020cross} proposed a method to combine cross-modality attention with semantic graph embedding. The method used an adjacency-based similarity graph embedding representation to learn semantic embeddings of labels.
Chen et al.~\cite{chen2022label} proposed a hierarchical residual network for hierarchical multi-granularity classification. In this network, knowledge between different layers can be transferred through residual structure, and the tree-structured information was aggregated using the proposed combined loss to train the model.

\begin{figure}[b]
\begin{center}
\includegraphics[width=1\linewidth]{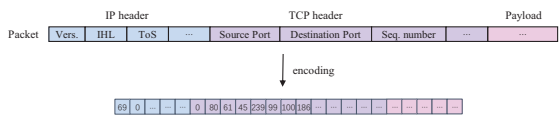}
\end{center}
   \caption{\emph{Byte encoding can map protocol fields in a packet to decimal numbers, turning the packet into a one-dimensional vector.}}
\label{fig:encoding}
\end{figure}

\begin{figure*}[t]
\begin{center}
\includegraphics[width=0.9\linewidth,keepaspectratio]{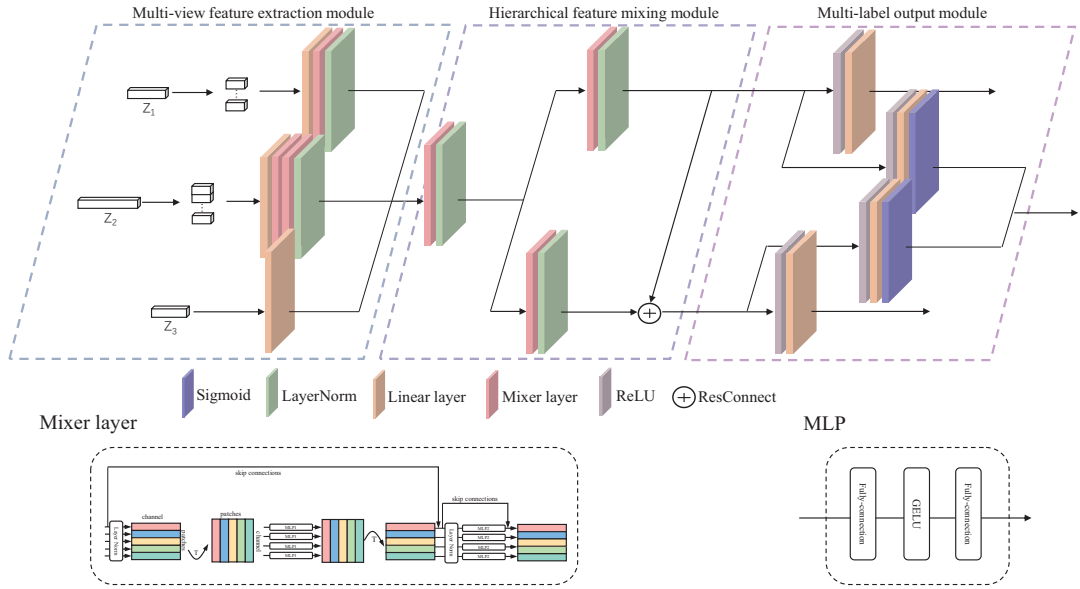}
\end{center}
   \caption{\emph{The framework of the network. The network structure consists of the multi-view feature extraction module, the hierarchical feature mixing module and the multi-label output module.  After the different features extracted from the three views are mixed, they are processed by specific-label levels. Then the child features inherit the attributes of the parent through the residual structure. Finally, three results will be output, from top to bottom are the parent prediction result, the hierarchical structure prediction result, and the child prediction result. The illustration of the Mixer layer comes from \cite{tolstikhin2021mlp}.
}}
\label{fig:model}
\end{figure*}

\section{Proposed Methods}
In this paper, we propose a MLP-Mixer based method, which combines multi-view, and multi-label technologies, and is named the multi-view multi-label traffic classification method. In this section, first, we introduce the encoding method used in our classification to process traffic packets. There are few detailed discussions on byte encoding in previous traffic classification studies. Second, our proposed method shows the multi-view feature extraction module, hierarchical feature mixing module, and multi-label output module.
The Mixer structure fits better with byte encoding, and is suitable for our proposed method. To the best of our knowledge, there is no existing research that applies the Mixer structure to traffic classification. Third, we describe the combinatorial loss used in our network. This combined loss can constrain the semantic hierarchy between multiple labels, which is beneficial for shared information between labels. This combined loss leverages the relationship between multi-labels from the hierarchical structure in the model.

\subsection{Data Encoding}
For data types with underlying semantics such as traffic data, the understanding of underlying data semantics by neural networks is the
foundation of classification tasks. A good coding approach can not only help the model understand the semantics, but also enhance the interpretability of the model. Byte encoding is employed in our method. This is a representation that encodes a single packet into a one-dimensional vector. With this encoding method, we can include the complete information of the data packet in the representation without the manual feature selection using domain expertise. As far as we know, the neural network has strong ability to extract features from the raw data packets~\cite{rezaei2019deep}. Retaining all the information can greatly save the time of manual feature extraction and reduce the requirement for professional knowledge.

Neural network requires fixed-size inputs. Our encoding method can truncate and pad the original packet to a fixed length based on prior knowledge of the length in different scenarios. Since our method needs to leverage the header and body of the data packet, the byte encoding can reduce the length of the input data as much as possible, compared with other encoding methods, such as binary encoding. Our method requires flow features of packets in addition to raw packets. The flow features of packets are usually in decimal. In order to ensure better network performance, we use byte encoding to convert the original packets into decimal to ensure the unity of network input. Normalizing the input of the network can greatly improve the performance of the network, which can not only reduce the training time but also improve the stability of the model~\cite{nayak2014impact}. The schematic diagram of byte encoding is showed in Fig.~\ref{fig:encoding}.

\subsection{Network Architecture}
Our network structure consists of three modules, multi-view feature extraction module, hierarchical feature mixing module, and multi-label output module, as showed in Fig.~\ref{fig:model}.
\subsubsection{Multi-view feature extraction module}
For network data packets, the header of the network protocol and its payload can be divided into two parts according to the OSI model. For the divided protocol headers, there are often more specific semantic structures and higher effective information concentration; for the divided payload, due to the diversity of upper-layer protocols and the encrypted payload, the information that the payload can provide for classification is not as effective and straight as the header part. In conventional methods, the header and payload are usually considered as a whole, but this type does not consider their different effects on the classifier. In our method, the packet fed into the model is a one-dimensional vector $Z$, $Z \in R^{n}$.
The header and the payload are divided according to the transport layer in the OSI model. Protocol headers at the transport layer and below tend to be fixed and usually not encrypted. The header of the transport layer and below is divided into the header part, the payload is divided into the payload part. According to these, we divide $Z$ into $Z_{1}$ and $Z_{2}$, corresponding to the header and payload respectively, and the two parts are sent to the network in parallel.

For the header part $Z_{1}$, we use a single Mixer layer for the feature extraction. Before passing through the Mixer layer, $Z_{1}$ is firstly divided into $S$ patches sequence, each patch length is $P$. The patch length $P$ is then mapped to the hidden dimension $C$ through a linear layer. The input table of $X \in R^{S*C}$ is obtained, and then sent to a single Mixer layer for feature extraction.
For header sections with well-defined semantic structure, the model's ability to understand the underlying data largely determines the model's performance. If features are extracted using convolution operations, details in mismatched local patterns may be ignored. Because the convolution operation is mainly used to extract the local pattern of the image, it acts on the underlying pixels without explicit semantics. Therefore, we use Mixer composed of MLP to extract header part feature information. In addition, we do not need a complex network to extraction for the header part due to the existence of specific semantics. So we adopt a single Mixer layer. 
For the payload part $Z_{2}$, we use two Mixer layers for the feature extraction. Also before feeding into the Mixer layers, the patch division and the mapping of the linear layer are passed. We hope to have a stronger representation ability to extract feature information for complex payload parts than the header.

In~\cite{tolstikhin2021mlp}, the Mixer layer is divided into two parts: token mixing and channel mixing. These correspond to different types of MLP layers for spatial location mixing and channel feature mixing respectively.
In our method, network traffic packets are one-dimensional structures, and we determine the patch size. 
For example, the patch  size is 20 bytes corresponding to the basic header length of the IP protocol and the TCP protocol.
We divide the header and payload parts based on the patch size, and the number of patches is determined by the length of the preprocessed packet and the patch size. 
After determining the patch size, a patch corresponds to a protocol header or the divided payload parts, while a channel corresponds to different protocol headers or different payload parts.
Compared to using a simple MLP structure, firstly, the Mixer layer needs to divide the input into patches, by adjusting the size of the patch, the patch can correspond to the protocol header in the packet. Secondly, MLP Mixers interleaves extracting the feature information of space and channel, corresponding to mixing inter-protocol information and intra-protocol information in packets.

In addition to considering the two views of packets, we also input the traffic flow features $Z_{3}$ to the model in parallel after statistics. For the task of traffic classification, the traffic flow in which the data packet is located can often provide sufficient classification information besides the information contained in a single data packet. For our network, there is a strong ability to extract the feature of the packet itself, but computing the flow features of the input data is not the field that our network is good at. To make up for this part, we can implement the flow feature statistic of input data packets through software and feed this part into the network in parallel. We wish our model to consider the features of the packet itself and the features of the flow to which the packet belongs.
\subsubsection{Hierarchical feature mixing module}
This module consists of three Mixer layers. The three features obtained by the multi-view feature extraction module are concatenated and fed into the first Mixer layer to communicate information from different views. The resulting output is then fed into two Mixer layers in parallel for feature extraction at the respective label level. The three Mixer layers have the same structure, the difference is that the two parallel Mixer layers are followed by the dimensionality reduction operation to compress the information between patches to facilitate the subsequent multi-label output module output results.
The first Mixer layer can maximize the mutual information between different views of the input data, resulting in a more compact and better feature representation. The mixed feature can consider three views simultaneously and contain the feature information of this category for the model to classify, which is the basis for the subsequent specific-label classification.
The next two Mixer layers are designed to extract specialized features at different label levels. Our model is used for multi-label classification, and the classification task levels corresponding to different labels need to extract different hierarchical features. In our multi-label classification task, a classification task consists of different related labels. The child label is obtained by the more fine-grained classification of the parent label, and the child label inevitably inherits the corresponding attributes of the parent label. That is to say, the features of the parent-child hierarchy also have an inheritance relationship, which is reflected by the residual structure of the two Mixer layers. The residual structure implements feature propagation across different labels, establishing connections between label-specific branches. The two Mixer layers extract the unique features of the corresponding label levels and merge the features extracted by the parent with the fine-grained features of the child through the residual structure, which reflects that the child class not only has its unique attributes but also inherits from the parent. 

\subsubsection{Multi-label output module}
The multi-label output module can output the classification results at different classification granularities. The classification labels of different scenes in the same dataset are correlated. We use a model to output the labels of several scenarios. The multi-label output module can output the classification results of all scenarios in parallel according to the hierarchical structure established between different scenarios, for example, VPN or non-VPN and chat class or email class. We only need to train once to meet the requirements of all scenarios in the hierarchical structure, which can greatly save us training models and adjusting parameter time. We hope that model training can also leverage the potential relationship between scenarios.

Three output channels are set up in the output module. The inputs of the two channels come from different levels of features of the hierarchical feature mixing module, both of which are composed of ReLU and linear layers, but correspond to classifications on different label categories.
The input of another channel is the features of two levels in the hierarchical feature mixing module, and then the output vectors of different categories are obtained through the ReLU, linear and sigmoid layers respectively.
By concatenating these two vectors to obtain a hierarchical structure vector, it can not only represent the category relationship of different levels, but also represent the inheritance relationship between the levels.
The output of the first channel is the coarse-grained parent label classification result, which is used to calculate the cross-entropy loss of the parent label. The second channel output contains both parent and child label hierarchy vectors, which are used to constrain hierarchical relationships. The hierarchical structure vector is output by different levels of sigmoid activation functions and then oncatenated to represent the inheritance and constraint relationship between multi-labels. Through the sigmoid activation function, each dimension of the hierarchical structure vector represents the probability of belonging to a category. The third channel outputs fine-grained child label classification results, which are used to calculate the cross-entropy loss of the child label.

\subsection{Loss Function}
Our combined loss consists of the parent-level classification loss, the child-level classification loss, and the hierarchical constraint loss. In the two different levels of parent and child, their corresponding labels focus on the classification of their respective levels, so we use cross-entropy for classification at different classification granularities which can strengthen the mutual exclusion relationship between different classes within the same level. In terms of the entire hierarchy, the semantic transfer relationship between the levels is also our focus in addition to the inside of the same level. But only using cross-entropy cannot satisfy the constraints on the hierarchy, so we introduce BCE (Binary Cross Entropy). We concatenate category labels of different granularities to form hierarchical structure labels. 
Hierarchical labels can not only describe the labels at each classification granularity in the dataset, but also describe the potential relationship between labels at different classification granularities, that is, the inheritance relationship between parent-child hierarchies. 
We calculate the BCE loss with the hierarchical structure vector obtained by the multi-label output module and the hierarchical structure label, which requires the model to pay attention to the relationship between the levels while paying attention to the internal classification of each level.
The dimensions inside the hierarchical structure vector are independent of each other. Each dimension represents the probability of belonging to each class and describes the hierarchical structure well. We combine the three losses by weights. This enables the model to pay attention to the inheritance relationship between categories while paying attention to different granularity classifications.
The weights can shift the focus of the model between different granular categories and inheritance relationships.
Our combinatorial loss is as follows: 
\begin{equation}
Loss=a*L_{parent}+b*L_{child}+c*L_{hierarchy}
\end{equation}
$a$ , $b$ , $c$ indicates the weights of the three losses, $L_{parent}$ and $L_{child}$ are the cross-entropy losses, and $ L_{hierarchy}$ is the BCE loss.

\section{Experiments}
\subsection{Dataset}
In previous works on network traffic classification, self-collected private traffic is often used, and traffic datasets are not published due to non-disclosure agreements, which seriously damage the credibility and reproducibility of their experimental results.
Our work selects three publicly available network traffic datasets, which are frequently used in network traffic classification. Two of the datasets are encrypted traffic of various applications, and the other dataset is from normal and malicious traffic. Since the input of our network needs to use the original packets and the flow features of the packets, the datasets we choose contain the original packets. At the same time, we use ISCXFlowMeter~\cite{lashkari2017characterization} , a flow feature extraction software written in JAVA, to extract features on the original packets to obtain the flow features we need. The three datasets are described as follow.

\subsubsection{Tor and non-Tor Dataset}
We use the "UNB-CIC Tor-non Tor" dataset, collected by Lashkari et al \cite{lashkari2017characterization}. The author used two virtual machines of the Whonix system to form the gateway and workstation, configured to route traffic through the Tor network, all Tor traffic must pass through the gateway to reach the workstation. Tor traffic packets were captured at the gateway, and non-Tor packets were captured at the workstation. These packets were collected in pcap file format. On the one hand, the pcap files can be differentiated according to the application that generated these packets (e.g. Facebook, Skype, etc.), and on the other hand, they can be divided according to the activities that these packets participated in (e.g. chat, file transfer, etc.).

\subsubsection{VPN and non-VPN Dataset}
We use the "ISCX VPN-non VPN" dataset, collected by Draper-Gil et al \cite{draper2016characterization}. This dataset was generated by the authors capturing daily traffic in the lab, where the captured packets were stored in pcap file format. These files are differentiated on the one hand by the application that generated the packets (e.g. Skype, Hangouts, etc.) and on the other hand, by the specific activities the packets were involved in (e.g. voice calls, chats, file transfers, and video calls). 
The dataset also includes packets captured through virtual private network sessions (VPN). As with non-VPN traffic, the capture of VPN traffic is also labeled by application and participating activities.

\subsubsection{USTC-TFC2016 Dataset}
We use the "USTC-TFC2016" dataset, collected by Wei et al.~\cite{wang2017malware}, which is a dataset of normal traffic and malware traffic. The part of normal traffic is 10 kinds of application traffic collected by IXIA BPS, IXIA BPS is a network traffic emulation device, these ten kinds of applications can be divided into 8 categories according to the specific activities involved. Malware traffic is derived in part from malware traffic from public websites collected by CTU researchers from 2011 to 2015.
The authors intercepted the overly large traffic and merged the too-small traffic, obtaining 10 types of malware traffic.

\subsection{ Pre-processing}
\subsubsection{Data Preprocessing}
As far as we know, original data structures like packets cannot be fed into the network by simple processing, and how to process such data is a critical step. A series of reasonable and effective processing steps will transform the underlying data into a network-friendly data form, which can not only contain the features of the data itself but also make the network easy to learn. There are three stages in our data preprocessing process, traffic cleaning, traffic encoding, and traffic format conversion.

\textbf{\emph{Step 1 (Traffic clear)}}.
This step is divided into packet cleaning and packet content cleaning. Packet cleaning aims to discard packets that are not related to the label class. Because the dataset we used was collected in the real network environment, there will be some packets that do not provide any information for classification. The dataset contains Domain Name System (DNS) resolution service  packets. DNS packets are used to resolve domain names to host addresses. This process does not bring useful information for distinguishing application activity, so we drop packets of this type. Since the TCP protocol has processes such as establishing a connection, terminating a connection, and acknowledging, packets without payloads are included in these processes. These packets participate in the TCP connection but have nothing to do with actual application activity, so we discard this type of data as well. The Address Resolution Protocol (ARP) resolves the IP address to obtain the MAC address and only provides support for the transmission of packets. We also discard such packets.

Packet content cleaning aims to deal with the structure inside the packet. Because the dataset collected complete packets, the packets often contained protocol headers at the data link layer, such as Ethernet headers. This protocol is more concerned with the information of the underlying physical link and has little to do with application activity, so we removed the Ethernet header from the packet. In the IP protocol of the network layer, there are two fields: destination IP and source IP. When different application activities occur between different IP addresses, there is a high probability that the IP addresses are used by the model to distinguish the application activities, which can be considered as cheating. We want the model to learn the essential differences between applications, so we anonymize the IP address field of the network layer.

\textbf{\emph{Step 2 (Traffic encoding)}}.
Neural networks require the input length to be consistent, and we need to unify the length of all packets. Considering that the maximum transmission length of standard Ethernet is 1500 bytes, we unified the length of all packets into 1500 bytes. Packets longer than 1500 bytes are truncated, and packets shorter than 1500 bytes are padded with 0 until their length is 1500 bytes. We use the proposed byte encoding to encode individual packets in the dataset into a one-dimensional vector. There are two protocols at the transport layer with different header lengths, TCP protocol and UDP protocol. In order to ensure the unity of the structure, we pad the UDP protocol header with 0 to make its length consistent with the length of the TCP protocol header. We use min-max normalization to normalize the vector between 0 and 1. In addition to the packets themselves, we use the ISCXFlowMeter flow feature extraction software to perform feature extraction on the flow. We select 8 useful flow features, "Flow Duration","Total Fwd Packet", "Total Bwd packets", "Flow Packets/s", "Flow IAT Mean", "Packet Length Mean", "Packet Length Std", "Average Packet Size", and convert them into one-dimensional vectors. We normalize each dimension inside the vector.

\textbf{\emph{Step 3 (CSV conversion)}}.
For the processed packet and flow features, we combine them with the corresponding label into one item. It is then stored as a CSV file.

\subsubsection{Labeling Dataset} 
The number of samples in the three datasets is very large and the data distribution is extremely imbalanced. 
In~\cite{wang2017malware,lotfollahi2020deep}, the relatively balanced datasets are used for experiments. 
We use the sampling technique to make the categories in each dataset as balanced as possible~\cite{lotfollahi2020deep}, and 
we can also compare the results fairly using the model proposed in \cite{lotfollahi2020deep}. We generate balanced three datasets by randomly sampling up 12,000 samples from each fine-grained category in the three datasets. If the sample size is not enough, we take all the data samples. 
The training set, validation set, and test set are divided from these datasets respectively. We randomly sample and divide each dataset into three subsets. The first is the training set, which accounts for 68$\%$ of the total number of samples, and is mainly used to train the model and adjust the parameters. The second is the validation set, which accounts for 16$\%$ of the total number of samples, and is mainly used to verify the effect of the model. The third is the test set, which accounts for 16$\%$ of the total number of samples, and is mainly used to test the model.

There are two scenarios considered in our classification task, according to which we need to assign multi-labels to three different datasets. In the first scenario, we first assign anomaly-level labels to each dataset, which is a binary classification task. For VPN and non-VPN, Tor and non-Tor datasets, these can be assigned anomaly detection labels by encrypted and non-encrypted. Encrypted packets are abnormal data that need to be detected from the perspective of anomaly detection, while non-encrypted packets are normal data. For the USTC dataset, the dataset can be assigned anomaly detection labels according to benign traffic and malicious traffic. Malicious traffic is abnormal traffic that we need to detect, and benign traffic is normal data. In the second scenario, we assign finer-grained labels to the three datasets based on the activities they were engaged in.

\begin{table*}
\centering
\caption{Anomaly label classification of balanced Tor and non-Tor dataset}
\label{Table:Top-Tor}
\resizebox{0.9\textwidth}{!}{
\begin{tabular}{|l|ccc|ccc||ccc|ccc|} 
\hline
\multirow{2}{*}{Class name} & \multicolumn{3}{c|}{1D-CNN}                                & \multicolumn{3}{c||}{Mixer}                                                 & \multicolumn{3}{c|}{Multi-view Multi-label MLP}                                                 & \multicolumn{3}{c|}{Multi-view Multi-label Mixer}                            \\ 
\cline{2-13}
                            & Pr                      & Re                      & F1     & Pr                      & Re                      & F1                      & Pr                      & Re                      & F1                      & Pr                      & Re                      & F1                       \\ 
\hline
non-Tor                     & 0.9996                  & \textbf{1.0000} & 0.9998 & 0.9997                  & \textbf{1.0000} & 0.9998                  & 0.9999                  & 0.9999                  & 0.9999                  & 0.9999                  & 0.9999                  & \textbf{0.9999}  \\
Tor                         & \textbf{1.0000} & 0.9996                  & 0.9998 & \textbf{1.0000} & 0.9997                  & \textbf{0.9999} & 0.9998                  & 0.9998                  & 0.9998                  & 0.9999                  & 0.9999                  & \textbf{0.9999}  \\ 
\hline
AVG                         & 0.9998                  & 0.9998                  & 0.9998 & 0.9998                  & \textbf{0.9999} & \textbf{0.9999} & \textbf{0.9999} & \textbf{0.9999} & \textbf{0.9999} & \textbf{0.9999} & \textbf{0.9999} & \textbf{0.9999}  \\
\hline
\end{tabular}
}
\end{table*}

\begin{table*}
\centering
\caption{Anomaly label classification of balanced VPN and non-VPN dataset}
\label{Table:Top-VPN}
\resizebox{0.9\textwidth}{!}{
\begin{tabular}{|l|ccc|ccc||ccc|ccc|} 
\hline
\multirow{2}{*}{Class name} & \multicolumn{3}{c|}{1D-CNN} & \multicolumn{3}{c||}{Mixer} & \multicolumn{3}{c|}{Multi-view Multi-label MLP}               & \multicolumn{3}{c|}{Multi-view Multi-label Mixer~}                           \\ 
\cline{2-13}
                            & Pr     & Re     & F1        & Pr     & Re     & F1        & Pr                      & Re     & F1     & Pr                      & Re                      & F1                       \\ 
\hline
non-VPN                     & 0.9944 & 0.9892 & 0.9918    & 0.9945 & 0.9923 & 0.9934    & \textbf{0.9982} & 0.9909 & 0.9945 & 0.9965                  & \textbf{0.9959} & \textbf{0.9962}  \\
VPN                         & 0.9894 & 0.9945 & 0.9919    & 0.9924 & 0.9946 & 0.9935    & 0.9735                  & 0.9948 & 0.9841 & \textbf{0.9959} & \textbf{0.9965} & \textbf{0.9962}  \\ 
\hline
AVG                         & 0.9919 & 0.9918 & 0.9918    & 0.9935 & 0.9934 & 0.9934    & 0.9859                  & 0.9928 & 0.9893 & \textbf{0.9962} & \textbf{0.9962} & \textbf{0.9962}  \\
\hline
\end{tabular}
}
\end{table*}

\begin{table*}
\centering
\caption{Anomaly label classification of balanced USTC-TFC2016 dataset}
\label{Table:Top-USTC}
\resizebox{0.9\textwidth}{!}{
\begin{tabular}{|l|ccc|ccc||ccc|ccc|} 
\hline
\multirow{2}{*}{Class name} & \multicolumn{3}{c|}{1D-CNN}                                                 & \multicolumn{3}{c||}{Mixer}                                & \multicolumn{3}{c|}{Multi-view Multi-label MLP} & \multicolumn{3}{c|}{Multi-view Multi-label Mixer~}          \\ 
\cline{2-13}
                            & Pr                      & Re                      & F1                      & Pr                      & Re                      & F1     & Pr     & Re     & F1     & Pr                      & Re                      & F1      \\ 
\hline
Benign                      & \textbf{1.0000} & \textbf{1.0000} & \textbf{1.0000} & 0.9998                  & \textbf{1.0000} & 0.9999 & 0.9998 & 0.9998 & 0.9998 & 0.9997                  & \textbf{1.0000} & 0.9998  \\
Malware                     & \textbf{1.0000} & \textbf{1.0000} & \textbf{1.0000} & \textbf{1.0000} & 0.9999                  & 0.9999 & 0.9997 & 0.9997 & 0.9997 & \textbf{1.0000} & 0.9998                  & 0.9999  \\ 
\hline
AVG                         & \textbf{1.0000} & \textbf{1.0000} & \textbf{1.0000} & 0.9999                  & 0.9999                  & 0.9999 & 0.9997 & 0.9997 & 0.9997 & 0.9998                  & 0.9999                  & 0.9999  \\
\hline
\end{tabular}
}
\end{table*}

\begin{table*}
\centering
\caption{Traffic label classification of balanced Tor and non-Tor dataset}
\label{Table:Traffic-Tor}
\resizebox{0.9\textwidth}{!}{
\begin{tabular}{|l|ccc|ccc||ccc|ccc|} 
\hline
\multirow{2}{*}{Class name} & \multicolumn{3}{c|}{1D-CNN}                                                 & \multicolumn{3}{c||}{Mixer}                                                 & \multicolumn{3}{c|}{Multi-view Multi-label MLP}                             & \multicolumn{3}{c|}{Multi-view Multi-label Mixer~}                           \\ 
\cline{2-13}
                            & Pr                      & Re                      & F1                      & Pr                      & Re                      & F1                      & Pr                      & Re                      & F1                      & Pr                      & Re                      & F1                       \\ 
\hline
Chat                        & 0.9106                  & 0.9303                  & 0.9204                  & 0.9511                  & 0.9808                  & 0.9657                  & 0.9286                  & 0.9715                  & 0.9496                  & \textbf{0.9652} & \textbf{0.9814} & \textbf{0.9732}  \\
Tor-Chat                    & 0.9979                  & 0.9964                  & 0.9972                  & \textbf{0.9990} & \textbf{0.9990} & \textbf{0.9990} & 0.9901                  & 0.9983                  & 0.9942                  & \textbf{0.9990} & \textbf{0.9990} & \textbf{0.9990}  \\
Email                       & 0.9811                  & 0.9985                  & 0.9897                  & 0.9920                  & 0.9990                  & 0.9955                  & 0.9969                  & 0.9979                  & 0.9974                  & \textbf{0.9970} & \textbf{0.9995} & \textbf{0.9982}  \\
Tor-Email                   & \textbf{1.0000} & \textbf{1.0000} & \textbf{1.0000} & 0.9995                  & 1.0000                  & 0.9997                  & 0.9984                  & \textbf{1.0000} & 0.9992                  & 0.9995                  & \textbf{1.0000} & 0.9997                   \\
Streaming                   & 0.8086                  & 0.8462                  & 0.8270                  & 0.9711                  & \textbf{0.9780} & \textbf{0.9745} & 0.8736                  & 0.8520                  & 0.8627                  & \textbf{0.9743} & 0.9673                  & 0.9708                   \\
Tor-Streaming               & 0.9978                  & 0.9776                  & 0.9876                  & 0.9952                  & 0.9973                  & 0.9963                  & 0.9942                  & 0.9942                  & 0.9942                  & \textbf{1.0000} & \textbf{1.0000} & \textbf{1.0000}  \\
Browing                     & 0.8148                  & 0.7648                  & 0.7890                  & \textbf{0.9655} & 0.9381                  & 0.9516                  & 0.8418                  & 0.8392                  & 0.8405                  & 0.9555                  & \textbf{0.9515} & \textbf{0.9535}  \\
Tor-Browing                 & 0.9984                  & 0.9938                  & 0.9961                  & \textbf{0.9995} & 0.9969                  & 0.9982                  & 0.9927                  & 0.9927                  & 0.9927                  & \textbf{0.9995} & \textbf{0.9995} & \textbf{0.9995}  \\
FileTransfer                & 0.9923                  & 0.9969                  & 0.9946                  & 0.9974                  & 0.9979                  & 0.9977                  & 0.9985                  & 0.9954                  & 0.9969                  & \textbf{1.0000} & \textbf{0.9995} & \textbf{0.9997}  \\
Tor-FileTransfer            & 0.9974                  & \textbf{1.0000} & 0.9987                  & \textbf{1.0000} & 0.9995                  & 0.9997                  & \textbf{1.0000} & 0.9969                  & 0.9985                  & \textbf{1.0000} & \textbf{1.0000} & \textbf{1.0000}  \\
VoIP                        & 0.9815                  & 0.9648                  & 0.9731                  & \textbf{0.9871} & \textbf{0.9835} & \textbf{0.9853} & 0.9807                  & 0.9786                  & 0.9797                  & 0.9856                  & \textbf{0.9835} & 0.9845                   \\
Tor-VoIP                    & 0.9985                  & 0.9990                  & 0.9987                  & \textbf{1.0000} & \textbf{0.9995} & \textbf{0.9997} & \textbf{1.0000} & 0.9953                  & 0.9976                  & 0.9995                  & \textbf{0.9995} & 0.9995                   \\
P2P                         & 0.9910                  & 0.9821                  & 0.9865                  & \textbf{0.9947} & 0.9874                  & 0.9910                  & 0.9927                  & 0.9836                  & 0.9881                  & \textbf{0.9947} & \textbf{0.9916} & \textbf{0.9932}  \\
Tor-P2P                     & 0.9772                  & 0.9970                  & 0.9870                  & 0.9975                  & 0.9960                  & 0.9968                  & \textbf{1.0000} & \textbf{1.0000} & \textbf{1.0000} & \textbf{1.0000} & 0.9995                  & 0.9998                   \\ 
\hline
AVG                         & 0.9605                  & 0.9605                  & 0.9604                  & 0.9893                  & 0.9895                  & 0.9893                  & 0.9706                  & 0.9711                  & 0.9708                  & \textbf{0.9907} & \textbf{0.9908} & \textbf{0.9908}  \\
\hline
\end{tabular}
}
\end{table*}

\begin{table*}
\centering
\caption{Traffic label classification of balanced VPN and non-VPN dataset}
\label{Table:Traffic-VPN}
\resizebox{0.9\textwidth}{!}{
\begin{tabular}{|l|ccc|ccc||ccc|ccc|} 
\hline
\multirow{2}{*}{Class name} & \multicolumn{3}{c|}{1D-CNN}               & \multicolumn{3}{c||}{Mixer}                                & \multicolumn{3}{c|}{Multi-view Multi-label MLP}            & \multicolumn{3}{c|}{Multi-view Multi-label Mixer}                            \\ 
\cline{2-13}
                            & Pr                      & Re     & F1     & Pr                      & Re                      & F1     & Pr                      & Re                      & F1     & Pr                      & Re                      & F1                       \\ 
\hline
Chat                        & 0.8692                  & 0.7687 & 0.8158 & 0.8571                  & 0.8182                  & 0.8372 & 0.9161                  & 0.9323                  & 0.9241 & \textbf{0.9742} & \textbf{0.9643} & \textbf{0.9692}  \\
Email                       & 0.7396                  & 0.9136 & 0.8174 & 0.8185                  & 0.8641                  & 0.8407 & 0.9511                  & \textbf{0.9774} & 0.9641 & \textbf{0.9670} & 0.9750                  & \textbf{0.9710}  \\
FileTransfer                & 0.9838                  & 0.9374 & 0.9601 & 0.9509                  & 0.9548                  & 0.9529 & \textbf{0.9937} & 0.9629                  & 0.9781 & 0.9908                  & \textbf{0.9938} & \textbf{0.9923}  \\
Streaming                   & 0.9910                  & 0.9878 & 0.9894 & 0.9909                  & 0.9852                  & 0.9880 & 0.9911                  & 0.9753                  & 0.9831 & \textbf{0.9936} & \textbf{0.9915} & \textbf{0.9926}  \\
Torrent                     & \textbf{0.9957} & 0.9930 & 0.9944 & \textbf{0.9957} & 0.9930                  & 0.9944 & 0.9933                  & 0.9959                  & 0.9946 & \textbf{0.9957} & \textbf{0.9968} & \textbf{0.9963}  \\
VoIP                        & 0.9680                  & 0.8962 & 0.9307 & 0.9426                  & 0.9233                  & 0.9328 & 0.9692                  & 0.9353                  & 0.9520 & \textbf{0.9903} & \textbf{0.9887} & \textbf{0.9895}  \\
VPN-Chat                    & 0.9776                  & 0.9851 & 0.9813 & 0.9795                  & 0.9810                  & 0.9802 & 0.9311                  & 0.9719                  & 0.9511 & \textbf{0.9892} & \textbf{0.9928} & \textbf{0.9910}  \\
VPN-FileTransfer            & 0.9842                  & 0.9837 & 0.9840 & 0.9787                  & 0.9797                  & 0.9792 & 0.9536                  & 0.9809                  & 0.9670 & \textbf{0.9869} & \textbf{0.9919} & \textbf{0.9894}  \\
VPN-Email                   & 0.9867                  & 0.9952 & 0.9909 & 0.9815                  & 0.9968                  & 0.9891 & 0.9758                  & 0.9983                  & 0.9869 & \textbf{0.9957} & \textbf{0.9984} & \textbf{0.9971}  \\
VPN-Streaming               & 0.9975                  & 0.9894 & 0.9934 & 0.9975                  & 0.9904                  & 0.9939 & 0.9940                  & \textbf{0.9955} & 0.9947 & \textbf{0.9980} & 0.9944                  & \textbf{0.9962}  \\
VPN-Torrent                 & 0.9979                  & 0.9974 & 0.9977 & 0.9964                  & \textbf{0.9979} & 0.9972 & 0.9939                  & 0.9835                  & 0.9887 & \textbf{1.0000} & 0.9964                  & \textbf{0.9982}  \\
VPN-VoIP                    & 0.9840                  & 0.9840 & 0.9840 & 0.9867                  & 0.9883                  & 0.9875 & 0.9665                  & 0.9794                  & 0.9729 & \textbf{0.9952} & \textbf{0.9920} & \textbf{0.9936}  \\ 
\hline
AVG                         & 0.9563                  & 0.9526 & 0.9533 & 0.9563                  & 0.9561                  & 0.9561 & 0.9691                  & 0.9741                  & 0.9714 & \textbf{0.9897} & \textbf{0.9897} & \textbf{0.9897}  \\
\hline
\end{tabular}
}
\end{table*}

\begin{table*}
\centering
\caption{Traffic label classification of balanced USTC-TFC2016 dataset}
\label{Table:Traffic-USTC}
\resizebox{0.9\textwidth}{!}{
\begin{tabular}{|l|ccc|ccc||ccc|ccc|} 
\hline
\multirow{2}{*}{Class name} & \multicolumn{3}{c|}{1D-CNN}                                                 & \multicolumn{3}{c||}{Mixer}                                                 & \multicolumn{3}{c|}{Multi-view Multi-label MLP}                                                 & \multicolumn{3}{c|}{Multi-view Multi-label Mixer~}                           \\ 
\cline{2-13}
                            & Pr                      & Re                      & F1                      & Pr                      & Re                      & F1                      & Pr                      & Re                      & F1                      & Pr                      & Re                      & F1                       \\ 
\hline
Chat                        & \textbf{1.0000} & \textbf{1.0000} & \textbf{1.0000} & \textbf{1.0000} & \textbf{1.0000} & \textbf{1.0000} & \textbf{1.0000} & \textbf{1.0000} & \textbf{1.0000} & \textbf{1.0000} & \textbf{1.0000} & \textbf{1.0000}  \\
Dataset                     & \textbf{1.0000} & \textbf{1.0000} & \textbf{1.0000} & \textbf{1.0000} & \textbf{1.0000} & \textbf{1.0000} & 0.9995                  & \textbf{1.0000} & 0.9997                  & \textbf{1.0000} & \textbf{1.0000} & \textbf{1.0000}  \\
Email                       & 0.9995                  & 0.9989                  & 0.9992                  & 0.9979                  & 0.9989                  & 0.9984                  & 0.9995                  & \textbf{1.0000} & 0.9997                  & \textbf{1.0000} & \textbf{1.0000} & \textbf{1.0000}  \\
FileTransfer                & \textbf{0.9989} & \textbf{1.0000} & \textbf{0.9995} & 0.9984                  & 0.9984                  & 0.9984                  & 0.9984                  & 0.9984                  & 0.9984                  & \textbf{0.9989} & \textbf{1.0000} & \textbf{0.9995}  \\
Game                        & \textbf{1.0000} & \textbf{1.0000} & \textbf{1.0000} & 0.9995                  & \textbf{1.0000} & 0.9997                  & \textbf{1.0000} & 0.9995                  & 0.9997                  & \textbf{1.0000} & \textbf{1.0000} & \textbf{1.0000}  \\
P2P                         & 0.9975                  & 0.9992                  & 0.9983                  & 0.9983                  & 0.9992                  & 0.9987                  & \textbf{1.0000} & \textbf{1.0000} & \textbf{1.0000} & \textbf{1.0000} & \textbf{1.0000} & \textbf{1.0000}  \\
SocialNetwork               & 0.9995                  & \textbf{1.0000} & 0.9997                  & \textbf{1.0000} & \textbf{1.0000} & \textbf{1.0000} & \textbf{1.0000} & 0.9995                  & 0.9997                  & \textbf{1.0000} & \textbf{1.0000} & \textbf{1.0000}  \\
Streaming                   & \textbf{1.0000} & \textbf{1.0000} & \textbf{1.0000} & \textbf{1.0000} & \textbf{1.0000} & \textbf{1.0000} & \textbf{1.0000} & \textbf{1.0000} & \textbf{1.0000} & \textbf{1.0000} & \textbf{1.0000} & \textbf{1.0000}  \\
Cridex                      & 0.9995                  & \textbf{1.0000} & 0.9997                  & 0.9984                  & \textbf{1.0000} & 0.9992                  & \textbf{1.0000} & \textbf{1.0000} & \textbf{1.0000} & 0.9995                  & \textbf{1.0000} & 0.9997                   \\
Geodo                       & 0.9954                  & \textbf{0.9979} & \textbf{0.9966} & 0.9943                  & 0.9948                  & 0.9946                  & 0.9872                  & 0.9841                  & 0.9857                  & \textbf{0.9964} & 0.9969                  & \textbf{0.9966}  \\
Htbot                       & 0.9905                  & 0.9879                  & 0.9892                  & 0.9894                  & 0.9842                  & 0.9868                  & 0.9887                  & 0.9621                  & 0.9752                  & \textbf{0.9952} & \textbf{0.9910} & \textbf{0.9931}  \\
Miuref                      & \textbf{0.9953} & \textbf{1.0000} & \textbf{0.9976} & 0.9942                  & \textbf{1.0000} & 0.9971                  & 0.9771                  & \textbf{1.0000} & 0.9884                  & 0.9942                  & \textbf{1.0000} & 0.9971                   \\
Neris                       & 0.8439                  & 0.8213                  & 0.8324                  & 0.8524                  & 0.8680                  & 0.8601                  & 0.9534                  & 0.9463                  & 0.9498                  & \textbf{0.9581} & \textbf{0.9581} & \textbf{0.9581}  \\
Nsis-ay                     & 0.9776                  & 0.9872                  & 0.9824                  & 0.9831                  & 0.9866                  & 0.9849                  & 0.9921                  & 0.9751                  & 0.9835                  & \textbf{0.9938} & \textbf{0.9918} & \textbf{0.9928}  \\
Shifu                       & 0.9979                  & 0.9952                  & 0.9965                  & 0.9963                  & 0.9936                  & 0.9949                  & 0.9984                  & \textbf{0.9984} & \textbf{0.9984} & \textbf{0.9995} & 0.9973                  & \textbf{0.9984}  \\
Tinba                       & \textbf{1.0000} & \textbf{1.0000} & \textbf{1.0000} & \textbf{1.0000} & \textbf{1.0000} & \textbf{1.0000} & \textbf{1.0000} & \textbf{1.0000} & \textbf{1.0000} & \textbf{1.0000} & \textbf{1.0000} & \textbf{1.0000}  \\
Virut                       & 0.8413                  & 0.8477                  & 0.8445                  & 0.8795                  & 0.8601                  & 0.8697                  & 0.9308                  & 0.9538                  & 0.9422                  & \textbf{0.9631} & \textbf{0.9635} & \textbf{0.9633}  \\
Zeus                        & 0.9948                  & 0.9984                  & 0.9966                  & 0.9979                  & 0.9974                  & 0.9976                  & 0.9940                  & 0.9985                  & 0.9962                  & \textbf{1.0000} & \textbf{1.0000} & \textbf{1.0000}  \\ 
\hline
AVG                         & 0.9795                  & 0.9796                  & 0.9796                  & 0.9822                  & 0.9823                  & 0.9822                  & 0.9899                  & 0.9898                  & 0.9898                  & \textbf{0.9944} & \textbf{0.9944} & \textbf{0.9944}  \\
\hline
\end{tabular}
}
\end{table*}

\subsection{Implementaion Details}
The input of our network comes from three views, so we need to divide the 1*1500 data packet into 1*40 header part and 1*1460 body part, and feed the network together with 1*8 feature part. Our network is trained for 70 epochs. For the optimizer, we adopt SGD with momentum of 0.9 to optimize our model. The batch size is 256. The initial learning rate of the model is set to 0.1 and adjusted by the cosine annealing strategy. To prevent the model from overfitting, we set dropout to 0.3. 
There are also some parameters such as the number of Mixer layers for the header and payload. We determined these parameters through grid search and experiments.
\subsection{Evaluation Metrics}
In our experiments, we use three evaluation metrics to evaluate the performance of our model, i.e., precision (P), recall (R), and F1 score (F1). The precision, recall, and F1 score are used to indicate the classification performance of the model for each category.
\begin{equation}
    R_{i}=\frac{TP_{i}}{TP_{i}+TN_{i}} 
\end{equation}
\begin{equation}
    P_{i}=\frac{TP_{i}}{TP_{i}+FP_{i}} 
\end{equation}
\begin{equation}
	F1_{i}=\frac{2R_{i}P_{i}}{R_{i}+P_{i}}     
\end{equation}
$TP_{i}$ refers to the number of samples that the model correctly predicts from the $i$-th category to the $i$-th category; $FP_{i}$ refers to the number of samples that the model incorrectly predicts the samples that do not belong to the $i$-th category into the $i$-th category; $TN_{i}$ means that the model predicts the samples that do not belong to the $i$-th category into the number of samples that do not belong to the $i$-th category; $FN_{i}$ refers to the number of samples that belong to the $i$-th category wrongly predicted to be other categories.

\subsection{Compare With The Methods on Balanced Datasets} 
Our model has excellent performance on three public datasets. However, due to the differences in datasets, data processing methods, and experimental environments, it isn't easy to make a fair comparison between experimental results. Therefore, we will adopt the network structure commonly used in current traffic classification and show the superiority of our method in the traffic classification task through comparison. Below we will introduce the network structure of the three methods to compare with our model.

\textbf{1D-CNN}:
1D-CNN is a network that is often used in network traffic classification. It employs one-dimensional convolution operations to perform feature extraction on traffic data and then implements traffic classification through linear layers. Our 1D-CNN model comes from~\cite{lotfollahi2020deep}. Due to the different preprocessing and sampling methods of the traffic data, we do not directly use the results of that paper. Instead, we train the model in that paper on our dataset to achieve the best performance for comparison.

1D-CNN consists of two convolutional layers, one max-pooling layer, and three fully connected neural networks. The input to the 1D-CNN model is the entire data packet in one dimension. A two-dimensional tensor is obtained after the first two convolutional layers and the pooling layer. Then the two-dimensional tensor is flattened into a one-dimensional vector passed through three fully connected neural networks and finally used for classification tasks through softmax. To prevent overfitting, the dropout technique is employed with a dropout rate of 0.05. We adjust the category of the model output according to the task and use this model for the anomaly level classification and traffic level classification of the three datasets for a total of 6 experiments for comparison.

\textbf{Mixer}:
We also build a model based on the Mixer layer for traffic classification. This model consists of five simple layers, including four sequentially connected Mixer layers and one fully connected layer for mapping each category. Like the 1D-CNN model, the entire packet without division is fed into the network as input for classification. The output category of the model is set according to the classification task, and the model is used for classification experiments of anomaly detection labels and traffic labels on three datasets.

\textbf{Multi-view multi-label MLP}:
We set up a model multi-view multi-label MLP based on the MLP network structure for comparison. This model is consistent with our proposed model in the overall framework, including multi-view module, hierarchical feature module and multi-label module, but the Mixer layers in our framework is replaced by the MLP layers. The model is used to perform two classification experiments with different classification granularities on three datasets, and the comparison of the experimental results can show the advantages of the Mixer layer over the traditional MLP layer.

\subsubsection{Experimental Results}
In the anomaly detection label classification task, we can see that all four models achieve good performance, as indicated in Table \ref{Table:Top-Tor}, Table \ref{Table:Top-VPN}, Table \ref{Table:Top-USTC}. In both the Tor and VPN encryption datasets, the Mixer model outperforms 1D-CNN by a small margin. Multi-view multi-label MLP performs close to the Mixer model in TOR, but lags behind both in VPN. But all three models are slightly worse than ours. In the USTC dataset 1D-CNN leads by a slight advantage. The anomaly detection label classification task is essentially a binary classification task, and we can detect anomalies based on the classification results. The binary classification task is too simple for the experimental models, so the models have shown good results on the test set, and the difference in their experimental results cannot fully reflect the superiority of our model.
In the traffic label classification task, four models performed the multi-classification task, and the results are shown in Table \ref{Table:Traffic-Tor}, Table \ref{Table:Traffic-VPN}, Table \ref{Table:Traffic-USTC}. 
Our model shows absolute advantage over the other three models, and 1D-CNN lags behind the other three models. Compared with the multi-view multi-label MLP model, our model has the same framework, but the main structure of the model is changed from the MLP layer to the Mixer layer. From the experimental results, the effect of our model is better, which shows that the Mixer layer is more consistent with the protocol structure of network traffic packets, and can communicate more fully between and within the protocol. Compared with the Mixer model, our model also uses four advanced Mixer layers as the main structure of the network, but the frameworks of the two models are different. From the experimental results, our model performs better especially on the VPN dataset. This shows that the multi-view module designed by our method can provide more comprehensive information for the model, and at the same time, the transfer of features in the hierarchical feature module is more conducive to focusing on the features of the hierarchy itself and inheriting the features of the parent level.

\begin{table*}
\centering
\caption{Anomaly label classification of imbalanced Tor and non-Tor dataset}
\label{unbalanced anomaly:Tor}
\resizebox{0.9\textwidth}{!}{
\begin{tabular}{|l|ccc|ccc||ccc|ccc|} 
\hline
\multirow{2}{*}{Class name} & \multicolumn{3}{c|}{1D-CNN} & \multicolumn{3}{c||}{Mixer}                                                 & \multicolumn{3}{c|}{Multi-view Multi-label MLP}                                                 & \multicolumn{3}{c|}{Multi-view Multi-label Mixer}                            \\ 
\cline{2-13}
                            & Pr     & Re     & F1        & Pr                      & Re                      & F1                      & Pr                      & Re                      & F1                      & Pr                      & Re                      & F1                       \\ 
\hline
non-Tor                     & 0.9998 & 0.9998 & 0.9998    & \textbf{1.0000} & 0.9999                  & \textbf{1.0000} & 0.9999                  & \textbf{1.0000} & \textbf{1.0000} & \textbf{1.0000} & 0.9999                  & \textbf{1.0000}  \\
Tor                         & 0.9987 & 0.9991 & 0.9989    & 0.9996                  & \textbf{1.0000} & \textbf{0.9998} & \textbf{1.0000} & 0.9996                  & \textbf{0.9998} & 0.9996                  & \textbf{1.0000} & \textbf{0.9998}  \\ 
\hline
AVG                         & 0.9993 & 0.9994 & 0.9994    & 0.9998                  & \textbf{1.0000} & \textbf{0.9999} & \textbf{1.0000} & 0.9998                  & \textbf{0.9999} & 0.9998                  & \textbf{1.0000} & \textbf{0.9999}  \\
\hline
\end{tabular}
}
\end{table*}
\begin{table*}
\centering
\caption{Anomaly label classification of imbalanced VPN and non-VPN dataset}
\label{unbalanced anomaly:VPN}
\resizebox{0.9\textwidth}{!}{
\begin{tabular}{|l|ccc|ccc||ccc|ccc|} 
\hline
\multirow{2}{*}{Class name} & \multicolumn{3}{c|}{1D-CNN} & \multicolumn{3}{c||}{Mixer} & \multicolumn{3}{c|}{Multi-view Multi-label MLP} & \multicolumn{3}{c|}{Multi-view Multi-label Mixer}                            \\ 
\cline{2-13}
                            & Pr     & Re     & F1        & Pr     & Re     & F1        & Pr     & Re     & F1        & Pr                      & Re                      & F1                       \\ 
\hline
non-VPN                     & 0.9939 & 0.9939 & 0.9939    & 0.9941 & 0.9950 & 0.9945    & 0.9938 & 0.9928 & 0.9933    & \textbf{0.9977} & \textbf{0.9965} & \textbf{0.9971}  \\
VPN                         & 0.9639 & 0.9639 & 0.9639    & 0.9699 & 0.9649 & 0.9674    & 0.9574 & 0.9634 & 0.9604    & \textbf{0.9795} & \textbf{0.9866} & \textbf{0.9830}  \\ 
\hline
AVG                         & 0.9789 & 0.9789 & 0.9789    & 0.9820 & 0.9799 & 0.9810    & 0.9756 & 0.9781 & 0.9768    & \textbf{0.9886} & \textbf{0.9916} & \textbf{0.9901}  \\
\hline
\end{tabular}
}
\end{table*}
\begin{table*}
\centering
\caption{Anomaly label classification of imbalanced USTC-TFC2016 dataset}
\label{unbalanced anomaly:USTC}
\resizebox{0.9\textwidth}{!}{
\begin{tabular}{|l|ccc|ccc||ccc|ccc|} 
\hline
\multirow{2}{*}{Class name} & \multicolumn{3}{c|}{1D-CNN}                                                 & \multicolumn{3}{c||}{Mixer}                                & \multicolumn{3}{c|}{Multi-view Multi-label MLP} & \multicolumn{3}{c|}{Multi-view Multi-label Mixer}           \\ 
\cline{2-13}
                            & Pr                      & Re                      & F1                      & Pr                      & Re                      & F1     & Pr     & Re     & F1        & Pr                      & Re                      & F1      \\ 
\hline
Benign                      & \textbf{0.9998} & \textbf{1.0000} & \textbf{0.9999} & 0.9995                  & \textbf{1.0000} & 0.9998 & 0.9997 & 0.9999 & 0.9998    & \textbf{0.9998} & 0.9998                  & 0.9998  \\
Malware                     & \textbf{1.0000} & \textbf{0.9990} & \textbf{0.9995} & \textbf{1.0000} & 0.9979                  & 0.9990 & 0.9997 & 0.9986 & 0.9991    & 0.9993                  & \textbf{0.9990} & 0.9991  \\ 
\hline
AVG                         & \textbf{0.9999} & \textbf{0.9995} & \textbf{0.9997} & 0.9998                  & 0.9990                  & 0.9994 & 0.9997 & 0.9993 & 0.9995    & 0.9995                  & 0.9994                  & 0.9995  \\
\hline
\end{tabular}
}
\end{table*}
\begin{table*}
\centering
\caption{Traffic label classification of imbalanced Tor and non-Tor dataset}
\label{unbalanced traffic:Tor}
\resizebox{0.9\textwidth}{!}{
\begin{tabular}{|l|ccc|ccc||ccc|ccc|} 
\hline
\multirow{2}{*}{Class name} & \multicolumn{3}{c|}{1D-CNN}               & \multicolumn{3}{c||}{Mixer}                                                 & \multicolumn{3}{c|}{Multi-view Multi-label MLP}               & \multicolumn{3}{c|}{Multi-view Multi-label Mixer}                            \\ 
\cline{2-13}
                            & Pr                      & Re     & F1     & Pr                      & Re                      & F1                      & Pr     & Re                      & F1     & Pr                      & Re                      & F1                       \\ 
\hline
Chat                        & 0.9197                  & 0.9349 & 0.9272 & 0.9685                  & \textbf{0.9846} & \textbf{0.9765} & 0.9267 & 0.9092                  & 0.9178 & \textbf{0.9722} & 0.9794                  & 0.9758                   \\
Tor-Chat                    & 0.9847                  & 0.9699 & 0.9772 & 0.9910                  & \textbf{0.9970} & 0.9940                  & 0.9815 & 0.9608                  & 0.9711 & \textbf{1.0000} & 0.9940                  & \textbf{0.9970}  \\
Email                       & 0.9897                  & 0.9974 & 0.9935 & 0.9907                  & 0.9990                  & 0.9948                  & 0.9912 & 0.9943                  & 0.9927 & \textbf{0.9979} & \textbf{0.9995} & \textbf{0.9987}  \\
Tor-Email                   & \textbf{1.0000} & 0.9967 & 0.9983 & \textbf{1.0000} & \textbf{1.0000} & \textbf{1.0000} & 0.9967 & 0.9967                  & 0.9967 & \textbf{1.0000} & \textbf{1.0000} & \textbf{1.0000}  \\
Streaming                   & 0.8230                  & 0.8346 & 0.8288 & 0.9700                  & \textbf{0.9817} & 0.9758                  & 0.8238 & 0.8247                  & 0.8243 & \textbf{0.9796} & 0.9812                  & \textbf{0.9804}  \\
Tor-Streaming               & 0.9451                  & 0.9109 & 0.9277 & 0.9916                  & 0.9916                  & 0.9916                  & 0.9123 & 0.8691                  & 0.8902 & \textbf{0.9944} & \textbf{0.9972} & \textbf{0.9958}  \\
Browing                     & 0.8117                  & 0.7935 & 0.8025 & \textbf{0.9677} & 0.9369                  & 0.9521                  & 0.7863 & 0.8133                  & 0.7996 & 0.9648                  & \textbf{0.9578} & \textbf{0.9613}  \\
Tor-Browing                 & 0.9135                  & 0.9344 & 0.9238 & \textbf{0.9967} & 0.9902                  & 0.9934                  & 0.8540 & 0.9016                  & 0.8772 & \textbf{0.9967} & \textbf{1.0000} & \textbf{0.9984}  \\
FileTransfer                & 0.9897                  & 0.9984 & 0.9941 & 0.9969                  & 0.9979                  & 0.9974                  & 0.9990 & 0.9938                  & 0.9964 & \textbf{1.0000} & \textbf{0.9990} & \textbf{0.9995}  \\
Tor-FileTransfer            & 0.9755                  & 0.9876 & 0.9815 & \textbf{1.0000} & 0.9969                  & 0.9984                  & 0.9785 & 0.9876                  & 0.9831 & \textbf{1.0000} & \textbf{1.0000} & \textbf{1.0000}  \\
VoIP                        & 0.9891                  & 0.9660 & 0.9775 & 0.9868                  & 0.9905                  & 0.9886                  & 0.9766 & 0.9729                  & 0.9748 & \textbf{0.9899} & \textbf{0.9920} & \textbf{0.9910}  \\
Tor-VoIP                    & \textbf{1.0000} & 0.9935 & 0.9968 & 0.9968                  & 0.9968                  & 0.9968                  & 0.9808 & 0.9903                  & 0.9855 & 0.9968                  & \textbf{1.0000} & \textbf{0.9984}  \\
P2P                         & 0.9896                  & 0.9896 & 0.9896 & \textbf{0.9963} & 0.9876                  & 0.9919                  & 0.9895 & 0.9772                  & 0.9833 & 0.9953                  & \textbf{0.9912} & \textbf{0.9933}  \\
Tor-P2P                     & 0.9636                  & 0.9907 & 0.9770 & 0.9876                  & 0.9938                  & 0.9907                  & 0.9969 & \textbf{1.0000} & 0.9984 & \textbf{1.0000} & \textbf{1.0000} & \textbf{1.0000}  \\ 
\hline
AVG                         & 0.9496                  & 0.9499 & 0.9497 & 0.9886                  & 0.9889                  & 0.9887                  & 0.9424 & 0.9423                  & 0.9422 & \textbf{0.9920} & \textbf{0.9922} & \textbf{0.9921}  \\
\hline
\end{tabular}
}
\end{table*}
\begin{table*}
\centering
\caption{Traffic label classification of imbalanced VPN and non-VPN dataset}
\label{unbalanced traffic:VPN}
\resizebox{0.9\textwidth}{!}{
\begin{tabular}{|l|ccc|ccc||ccc|ccc|} 
\hline
\multirow{2}{*}{Class name} & \multicolumn{3}{c|}{1D-CNN}                                                 & \multicolumn{3}{c||}{Mixer}                                                 & \multicolumn{3}{c|}{Multi-view Multi-label MLP} & \multicolumn{3}{c|}{Multi-view Multi-label Mixer}                            \\ 
\cline{2-13}
                            & Pr                      & Re                      & F1                      & Pr                      & Re                      & F1                      & Pr     & Re     & F1        & Pr                      & Re                      & F1                       \\ 
\hline
Chat                        & 0.8973                  & 0.7721                  & 0.8300                  & 0.8793                  & 0.8320                  & 0.8550                  & 0.8881 & 0.9418 & 0.9142    & \textbf{0.9612} & \textbf{0.9654} & \textbf{0.9633}  \\
Email                       & 0.7594                  & 0.9359                  & 0.8385                  & 0.8433                  & 0.9008                  & 0.8711                  & 0.9498 & 0.9720 & 0.9608    & \textbf{0.9693} & \textbf{0.9791} & \textbf{0.9742}  \\
FileTransfer                & 0.9917                  & 0.9289                  & 0.9592                  & 0.9461                  & 0.9481                  & 0.9471                  & 0.9871 & 0.9533 & 0.9699    & \textbf{0.9926} & \textbf{0.9798} & \textbf{0.9862}  \\
Streaming                   & 0.9923                  & \textbf{0.9852} & 0.9888                  & 0.9892                  & 0.9796                  & 0.9844                  & 0.9800 & 0.9725 & 0.9762    & \textbf{0.9943} & 0.9842                  & \textbf{0.9892}  \\
Torrent                     & 0.9917                  & 0.9943                  & 0.9930                  & 0.9901                  & 0.9901                  & 0.9901                  & 0.9870 & 0.9875 & 0.9873    & \textbf{0.9958} & \textbf{0.9969} & \textbf{0.9964}  \\
VoIP                        & 0.9534                  & 0.9133                  & 0.9329                  & 0.9406                  & 0.9233                  & 0.9318                  & 0.9791 & 0.9343 & 0.9562    & \textbf{0.9821} & \textbf{0.9827} & \textbf{0.9824}  \\
VPN-Chat                    & 0.9105                  & 0.9105                  & 0.9105                  & 0.9083                  & 0.9167                  & 0.9124                  & 0.8634 & 0.8580 & 0.8607    & \textbf{0.9422} & \textbf{0.9568} & \textbf{0.9495}  \\
VPN-FileTransfer            & 0.9452                  & 0.9391                  & 0.9421                  & 0.9632                  & 0.9231                  & 0.9427                  & 0.8656 & 0.8462 & 0.8558    & \textbf{0.9712} & \textbf{0.9712} & \textbf{0.9712}  \\
VPN-Email                   & \textbf{0.9589} & \textbf{0.9849} & \textbf{0.9718} & 0.9231                  & 0.9759                  & 0.9488                  & 0.9412 & 0.9639 & 0.9524    & 0.9534                  & \textbf{0.9849} & 0.9689                   \\
VPN-Streaming               & 0.9782                  & \textbf{0.9968} & 0.9874                  & 0.9874                  & 0.9937                  & \textbf{0.9905} & 0.9480 & 0.9841 & 0.9657    & \textbf{0.9905} & 0.9905                  & \textbf{0.9905}  \\
VPN-Torrent                 & 0.9853                  & 0.9853                  & 0.9853                  & \textbf{0.9912} & \textbf{0.9941} & \textbf{0.9926} & 0.9006 & 0.9086 & 0.9046    & \textbf{0.9912} & \textbf{0.9941} & \textbf{0.9926}  \\
VPN-VoIP                    & 0.9421                  & 0.9272                  & 0.9346                  & 0.9340                  & 0.9399                  & 0.9369                  & 0.9681 & 0.9589 & 0.9634    & \textbf{0.9744} & \textbf{0.9652} & \textbf{0.9698}  \\ 
\hline
AVG                         & 0.9422                  & 0.9395                  & 0.9395                  & 0.9413                  & 0.9431                  & 0.9420                  & 0.9382 & 0.9401 & 0.9389    & \textbf{0.9765} & \textbf{0.9792} & \textbf{0.9778}  \\
\hline
\end{tabular}
}
\end{table*}

\begin{table*}
\centering
\caption{Traffic label classification of imbalanced USTC-TFC2016 dataset}
\label{unbalanced traffic:USTC}
\resizebox{0.9\textwidth}{!}{
\begin{tabular}{|l|ccc|ccc||ccc|ccc|} 
\hline
\multirow{2}{*}{Class name} & \multicolumn{3}{c|}{1D-CNN}                                                 & \multicolumn{3}{c||}{Mixer}                                                 & \multicolumn{3}{c|}{Multi-view Multi-label MLP}                                                    & \multicolumn{3}{c|}{Multi-view Multi-label Mixer}                            \\ 
\cline{2-13}
                            & Pr                      & Re                      & F1                      & Pr                      & Re                      & F1                      & Pr                      & Re                      & F1                      & Pr                      & Re                      & F1                       \\ 
\hline
Chat                        & \textbf{1.0000} & \textbf{1.0000} & \textbf{1.0000} & \textbf{1.0000} & \textbf{1.0000} & \textbf{1.0000} & \textbf{1.0000} & \textbf{1.0000} & \textbf{1.0000} & \textbf{1.0000} & \textbf{1.0000} & \textbf{1.0000}  \\
Dataset                     & \textbf{1.0000} & \textbf{1.0000} & \textbf{1.0000} & 0.9984                  & \textbf{1.0000} & 0.9992                  & 0.9989                  & 0.9989                  & 0.9989                  & \textbf{1.0000} & 0.9995                  & 0.9997                   \\
Email                       & 0.9985                  & 0.9995                  & 0.9990                  & 0.9990                  & \textbf{1.0000} & 0.9995                  & 0.9969                  & 1.0000                  & 0.9985                  & \textbf{0.9995} & 0.9995                  & \textbf{0.9995}  \\
FileTransfer                & 0.9979                  & \textbf{1.0000} & 0.9990                  & 0.9974                  & 0.9995                  & 0.9984                  & \textbf{0.9990} & 0.9979                  & 0.9984                  & 0.9984                  & \textbf{1.0000} & \textbf{0.9992}  \\
Game                        & \textbf{1.0000} & \textbf{1.0000} & \textbf{1.0000} & 0.9990                  & \textbf{1.0000} & 0.9995                  & \textbf{1.0000} & 0.9995                  & 0.9997                  & \textbf{1.0000} & \textbf{1.0000} & \textbf{1.0000}  \\
P2P                         & \textbf{1.0000} & 0.9975                  & 0.9987                  & 0.9992                  & \textbf{1.0000} & 0.9996                  & \textbf{1.0000} & 0.9992                  & \textbf{0.9996} & 0.9992                  & 0.9992                  & 0.9992                   \\
SocialNetwork               & 0.9984                  & \textbf{1.0000} & 0.9992                  & \textbf{1.0000} & \textbf{1.0000} & \textbf{1.0000} & 0.9995                  & 0.9989                  & 0.9992                  & 0.9989                  & \textbf{1.0000} & 0.9995                   \\
Streaming                   & 0.9990                  & \textbf{1.0000} & 0.9995                  & 0.9990                  & \textbf{1.0000} & 0.9995                  & \textbf{1.0000} & \textbf{1.0000} & \textbf{1.0000} & \textbf{1.0000} & \textbf{1.0000} & \textbf{1.0000}  \\
Cridex                      & 0.9669                  & 0.9907                  & 0.9787                  & 0.9425                  & 0.9105                  & 0.9262                  & \textbf{1.0000} & \textbf{1.0000} & \textbf{1.0000} & \textbf{1.0000} & \textbf{1.0000} & \textbf{1.0000}  \\
Geodo                       & 0.9773                  & 0.9321                  & 0.9542                  & 0.9012                  & 0.9290                  & 0.9149                  & 0.9497                  & 0.9321                  & 0.9408                  & \textbf{0.9816} & \textbf{0.9877} & \textbf{0.9846}  \\
Htbot                       & 0.9464                  & 0.9493                  & 0.9478                  & \textbf{0.9878} & 0.9672                  & \textbf{0.9774} & 0.9254                  & 0.9254                  & 0.9254                  & 0.9731                  & \textbf{0.9701} & 0.9716                   \\
Miuref                      & 0.9659                  & \textbf{0.9905} & 0.9781                  & 0.9720                  & \textbf{0.9905} & 0.9811                  & 0.9630                  & \textbf{0.9905} & 0.9765                  & \textbf{0.9873} & 0.9873                  & \textbf{0.9873}  \\
Neris                       & 0.8480                  & 0.8158                  & 0.8316                  & 0.8540                  & 0.8041                  & 0.8283                  & 0.9088                  & 0.8450                  & 0.8758                  & \textbf{0.9231} & \textbf{0.9123} & \textbf{0.9176}  \\
Nsis-ay                     & 0.9531                  & 0.9839                  & 0.9683                  & 0.9681                  & 0.9774                  & 0.9727                  & 0.9742                  & 0.9742                  & 0.9742                  & \textbf{0.9838} & \textbf{0.9806} & \textbf{0.9822}  \\
Shifu                       & 0.9871                  & \textbf{0.9968} & 0.9919                  & \textbf{0.9935} & 0.9935                  & 0.9935                  & 0.9903                  & \textbf{0.9968} & 0.9935                  & \textbf{0.9935} & \textbf{0.9968} & \textbf{0.9951}  \\
Tinba                       & \textbf{1.0000} & \textbf{1.0000} & \textbf{1.0000} & 0.9231                  & \textbf{1.0000} & 0.9600                  & \textbf{1.0000} & \textbf{1.0000} & \textbf{1.0000} & \textbf{1.0000} & \textbf{1.0000} & \textbf{1.0000}  \\
Virut                       & 0.8282                  & 0.8108                  & 0.8194                  & 0.8195                  & 0.8318                  & 0.8256                  & 0.8531                  & 0.9069                  & 0.8792                  & \textbf{0.9224} & \textbf{0.9279} & \textbf{0.9251}  \\
Zeus                        & 0.9744                  & 0.9682                  & 0.9712                  & 0.9936                  & \textbf{0.9936} & \textbf{0.9936} & 0.9839                  & 0.9745                  & 0.9792                  & \textbf{0.9968} & 0.9873                  & 0.9920                   \\ 
\hline
AVG                         & 0.9690                  & 0.9686                  & 0.9687                  & 0.9637                  & 0.9665                  & 0.9649                  & 0.9746                  & 0.9744                  & 0.9744                  & \textbf{0.9865} & \textbf{0.9860} & \textbf{0.9863}  \\
\hline
\end{tabular}
}
\end{table*}

\begin{figure}[h]
    \centering
    \includegraphics[width=0.8\linewidth]{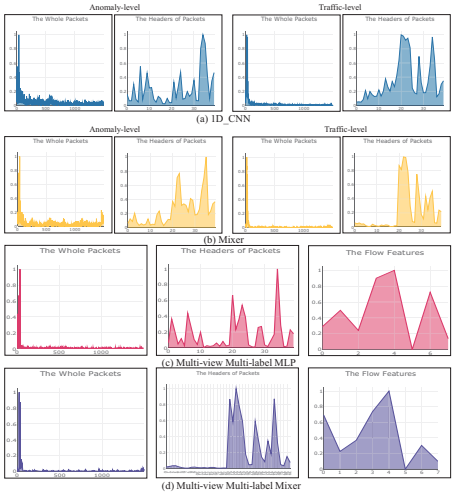}
    \caption{Input importance visualization of Tor and non-Tor dataset.}
    \label{fig:explain_Tor}
\end{figure}

\begin{figure}[h]
     \centering
     \includegraphics[width=0.8\linewidth]{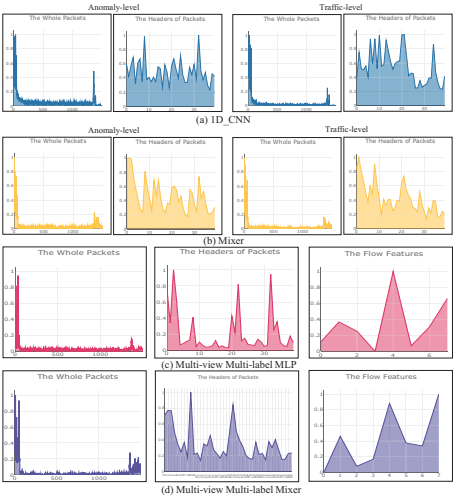}
     \caption{Input importance visualization of VPN and non-VPN dataset.}
     \label{fig:explain_VPN}
 \end{figure}

\begin{figure}[h]
     \centering
     \includegraphics[width=0.8\linewidth]{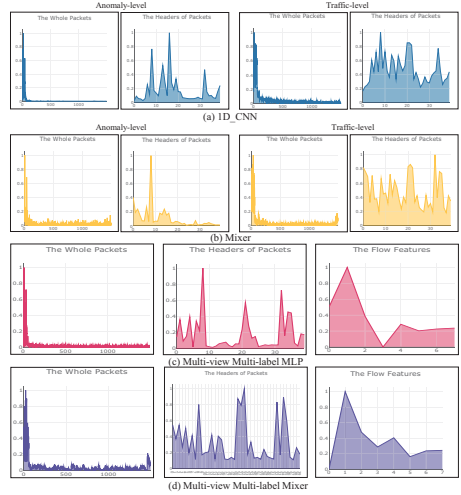}
     \caption{Input importance visualization of USTC-TFC2016 dataset.}
     \label{fig:explain_USTC}
     \vspace*{-0.3cm}
 \end{figure}

\subsubsection{Visualization Experiments}
We conduct visualization experiments to show how important the input is in different models. We hope to see a visual representation of the inputs' importance to better understand how our model works. As a result, we plot the importance of the four models on the all three datasets for a given number of input samples. The x-axis of each figure corresponds to the byte number, and the y-axis corresponds to the relative importance as showed in Fig.~\ref{fig:explain_Tor}, Fig.~\ref{fig:explain_VPN}, Fig.~\ref{fig:explain_USTC}. In the figures of the two comparison models, 1D-CNN and Mixer, the figures on the left of each label level correspond to the importance of the entire packet, and the figures on the right are the importance of the packet header, which are used to compare with our model. In addition to the importance of the whole packet and the the header part, there is also importance figure of the flow features in multi-view multi-label MLP and our model.
In the importance figure of multi-view multi-label MLP, the activation part of the whole packet and the packet header are much less than our method in the USTC and VPN datasets. But in the TOR dataset, multi-view multi-label MLP has more activation parts.

The following mainly analyzes the importance figures of our model.
From the figures of the importance of the whole packet, in the three datasets, the header part of the packet has a relatively large impact on the classification results, and the body part also has a impact on the results, especially in the USTC and VPN datasets. From the importance figures of the header part in the packet, the main roles of the Tor dataset include the source and destination ports of the TCP protocol (bytes 20 to 23), a portion of the TCP protocol sequence number (byte 24), a portion of the ACK number in the packet (bytes 28 to 29), and the window size in the TCP protocol (bytes 34 and 35). It should be noted that the TCP protocol header in the Tor data packet is more important for network classification, while the IP protocol header is relatively less important. The main roles of the VPN dataset include the version number and header length, service type, and total length of the IP packet (bytes 0 to 3), the TTL of the IP protocol (byte 8), source and destination port numbers of the TCP protocol (bytes 20 to 23), and the window size of the TCP protocol (byte 34). In our method, the TTL of the IP protocol in VPN packets is very helpful for classification. The main roles of USTC dataset include the IP protocol's version number and header length (byte 0), the total length of the IP protocol (byte 2), a portion of the identification in the IP protocol (byte 4), the flags in the IP protocol (byte 6), the TTL in the IP protocol (byte 8), the source and destination port numbers in the TCP protocol (bytes 20 to 22), the header length of the TCP protocol (byte 32), and the window size in the TCP protocol (bytes 34 and 35).
From the flow feature importance figures, 
it can be seen that in the Tor and VPN datasets, "Flow Packets/s" have high importance. In addition, "Average Packet Size" is also important in the VPN dataset. In the USTC dataset, "Total Fwd Packets" has high importance.

\subsection{Compare With The Methods on Imbalanced Datasets}
In the actual network environment, abnormal traffic is rare, so we change the number of abnormal categories to one-third of the original on the basis of balanced datasets so that we get imbalanced datasets on three datasets respectively. And divide the training set, verification set and test set according to the above settings. Similarly, we use three methods of 1D-CNN, Mixer and multi-view multi-label MLP for comparative experiments.

\subsubsection{Experimental Results}
As shown in the table \ref{unbalanced anomaly:Tor}, table \ref{unbalanced anomaly:VPN}, table \ref{unbalanced anomaly:USTC}, of the three anomaly label classification tasks, on the results of the TOR and VPN datasets, our model achieves better results than other models, but on the USTC dataset, 1D-cnn is still slightly lead.
As shown in the table \ref{unbalanced traffic:Tor}, table \ref{unbalanced traffic:VPN}, table \ref{unbalanced traffic:USTC}, our model achieves the best results on three imbalanced datasets.
When the datasets become imbalanced, the performance of all other models decrease significantly, but the decrease on our model is minimal and still remaines at a high level, and even improves slightly on the TOR dataset.
Our multi-view feature extraction module and hierarchical feature mixing module have strong feature extraction capabilities, which can extract targeted and effective features, and further integrate features through the residual structure.
At the same time, our loss function consists of three parts. During the training, the introduced hierarchical loss enables our model to focus on fewer anomalous samples.
Therefore, even if the number of anomalous samples decrease, our model can maintain relatively good performance.

\begin{table*}
\centering
\caption{Ablation experiment}
\label{Table:Ablation}
\resizebox{0.9\textwidth}{!}{
\begin{tabular}{|cl|ll|lcccc|cccc|} 
\hline
\multicolumn{1}{|l}{\multirow{2}{*}{Dataset}} &  & \multicolumn{1}{c}{\multirow{2}{*}{Model}} &  &  & \multicolumn{3}{c}{Anomaly-level} & \multicolumn{1}{l|}{} & \multicolumn{1}{l}{} & \multicolumn{3}{c|}{Traffic-level}  \\ 
\cline{5-13}
\multicolumn{1}{|l}{}                         &  & \multicolumn{1}{c}{}                       &  &  & Pr     & Re     & F1              &                       &                      & Pr     & Re     & F1                \\ 
\hline
\multirow{4}{*}{Tor}                          &  & w/o multi-view                             &  &  & 0.9999 & 0.9999 & 0.9999          &                       &                      & 0.9883 & 0.9886 & 0.9884            \\
                                              &  & w/o multi-label on anomaly level           &  &  & 0.9999 & 1      & 0.9999          &                       &                      & -      & -      & -                 \\
                                              &  & w/o multi-label on traffic level           &  &  & -      & -      & -               &                       &                      & 0.9897 & 0.9899 & 0.9898            \\ 
\cline{3-13}
                                              &  & our model                                  &  &  & 0.9999 & 0.9999 & 0.9999          &                       &                      & 0.9907 & 0.9908 & 0.9908            \\ 
\hline
\multirow{4}{*}{VPN}                          &  & w/o multi-view                             &  &  & 0.9938 & 0.9938 & 0.9938          &                       &                      & 0.9561 & 0.9549 & 0.9551            \\
                                              &  & w/o multi-label on anomaly level           &  &  & 0.9951 & 0.9952 & 0.9952          &                       &                      & -      & -      & -                 \\
                                              &  & w/o multi-label on traffic level           &  &  & -      & -      & -               &                       &                      & 0.9880 & 0.9880 & 0.9880            \\ 
\cline{3-13}
                                              &  & our model                                  &  &  & 0.9962 & 0.9962 & 0.9962          &                       &                      & 0.9897 & 0.9897 & 0.9897            \\ 
\hline
\multirow{4}{*}{USTC}                         &  & w/o multi-view                             &  &  & 0.9998 & 0.9998 & 0.9998          &                       &                      & 0.9777 & 0.9832 & 0.9802            \\
                                              &  & w/o multi-label on anomaly level           &  &  & 1      & 1      & 1               &                       &                      & -      & -      & -                 \\
                                              &  & w/o multi-label on traffic level           &  &  & -      & -      & -               &                       &                      & 0.9922 & 0.9922 & 0.9922            \\ 
\cline{3-13}
                                              &  & our model                                  &  &  & 0.9998 & 0.9999 & 0.9999          &                       &                      & 0.9944 & 0.9944 & 0.9944            \\
\hline
\end{tabular}
}
\end{table*}

\subsection{Ablation Study}
In this section, we will conduct ablation experiments on all three datasets. We will study two key parts of our model to demonstrate the roles of multi-view and multi-label.
\subsubsection{Significance of Multi-view}
Instead of splitting the packet into two parts, we feed the packet as a whole to the model and remove the view of flow features to demonstrate the role of multi-view. As the Table~\ref{Table:Ablation} shows, we can see that the model without multi-view lags slightly behind our model on the anomaly level classification of the three datasets. Because anomaly level is a simple task of binary classification, these models perform relatively well, so the gap between the models is not obvious. However, at the traffic level of the three datasets, the gap is more obvious, especially for the VPN and non-VPN datasets.

\subsubsection{Contribution of Multi-label}
We only use a single label for training and remove the hierarchical constraint loss to demonstrate the power of multi-label. As the Table~\ref{Table:Ablation} shows, We used two single-label models to train on the anomaly level and traffic level tasks. We can see that the model without multi-label performs almost the same as our model on the anomaly level tasks of the three datasets, all reaching a high level. But on traffic level tasks of three datasets, the model without multi-label lags behind our model. 

\section{Conclusion}
In this paper, we study the anomaly network traffic classification problem and propose a MLP-Mixer based neural network with multi-view and multi-label setting. We employ the MLP-Mixer operation as the building block of network. Compared with the convolution operation, the Mixer structure considers the global relationship of data packets to extract more discriminative features. In addition, we employ the multi-view module to provide diverse information of data packets for enhancing the performance of model. The multi-label module can make use of labels in multiple scenarios at the same time to improve the classification accuracy of all scenarios simultaneously. We embed these modules into our model to train an end-to-end network. Experiments are conducted on three public datasets, the experimental results have demonstrated that our method has superior performance comparing with existing network traffic classification methods.

\bibliographystyle{ieeetr}
\bibliography{Example.bbl}

\vfill
\end{document}